\documentclass[11pt]{article}

\usepackage[utf8]{inputenc}
\usepackage{amsmath, amssymb}
\usepackage{graphicx}
\usepackage{xcolor}
\usepackage{hyperref}
\usepackage{enumitem}
\usepackage{booktabs}  
\usepackage{threeparttable}  
\usepackage{tikz}
\usepackage{adjustbox}
\usepackage{pgfplots}
\pgfplotsset{compat=1.18}
\usetikzlibrary{positioning, fit, matrix, backgrounds, shapes.geometric, arrows.meta, patterns, pgfplots.groupplots, plotmarks}
\usepgfplotslibrary{polar}

\newcommand{\orcid}[1]{\href{https://orcid.org/#1}{\texttt{ORCID}}}

\tikzset{
  data/.style={
    rounded corners, fill=green!15, draw=black,
    minimum width=2.2cm, minimum height=0.9cm,
    align=center, font=\scriptsize, text width=2.1cm
  },
  process/.style={
    rounded corners, fill=blue!15, draw=black,
    minimum width=3.4cm, minimum height=0.9cm,
    align=center, font=\scriptsize, text width=2.7cm
  },
  output/.style={
    rounded corners, fill=red!15, draw=black,
    minimum width=1.4cm, minimum height=0.9cm,
    align=center, font=\bfseries\scriptsize, text width=3.5cm
  },
  arrow/.style={
    -{Stealth}, line width=0.5pt, shorten >=1pt, shorten <=1pt
  },
}

\title{From Extraction to Synthesis: \\ Entangled Heuristics for Agent-Augmented Strategic Reasoning}
\author{
  Renato Ghisellini\textsuperscript{1}\,\orcid{0000-0002-8461-8400} \\
  Remo Pareschi\textsuperscript{2}\thanks{Corresponding author: remo.pareschi@unimol.it}\,\orcid{0000-0002-4912-582X} \\
  Marco Pedroni\textsuperscript{1}\,\orcid{0009-0005-9516-4079} \\
  Giovanni Battista Raggi\textsuperscript{1}\,\orcid{0009-0004-2684-9464} \\\\
  \textsuperscript{1}Institute for Generative Strategy, Ferrara \\\\
  \textsuperscript{2}Stake Lab, University of Molise
}
\date{}

\begin{document}

\maketitle

\begin{abstract}
We present a hybrid architecture for agent-augmented strategic reasoning, combining heuristic extraction, semantic activation, and compositional synthesis. Drawing on sources ranging from classical military theory to contemporary corporate strategy, our model activates and composes multiple heuristics through a process of semantic interdependence inspired by research in quantum cognition. Unlike traditional decision engines that select the best rule, our system fuses conflicting heuristics into coherent and context-sensitive narratives, guided by semantic interaction modeling and rhetorical framing. We demonstrate the framework via a Meta vs. FTC case study, with preliminary validation through semantic metrics. Limitations and extensions (e.g., dynamic interference tuning) are discussed.
\end{abstract}

\section{Introduction}
The increasing deployment of artificial agents in domains involving strategic reasoning — such as policy design, competitive business analysis, and geopolitical simulation — raises an urgent challenge: How can we equip these agents with not only analytical capabilities but also access to the tacit, historically grounded heuristics that human strategists rely on?

Recent work in explainable and human-aligned AI has turned toward mining and applying heuristic knowledge~\cite{recommending_strategies} organized as conditional decision rules. Such prior research proposes an architecture that links analytical frameworks (e.g., SWOT~\cite{helms2010exploring}, Porter's Five Forces~\cite{porter2008five}) with culturally and historically embedded decision heuristics (e.g., the Thirty-Six Stratagems of Chinese military tradition~\cite{taylor2013thirty,von1991book}), using semantic inference to recommend plausible strategic actions based on situational profiles.

This paper extends that line of inquiry in a new direction. Rather than merely filtering and recommending discrete heuristics, we propose a generative synthesis engine that composes strategic reasoning across multiple, possibly conflicting axioms. Drawing on empirical findings from quantum cognition research, which documents systematic violations of classical logic in human reasoning, such as conjunction fallacies and order effects, we model strategic heuristics not as mutually exclusive options but as semantically interrelated potentials that can be composed into coherent formulations.

Throughout this paper, we employ the term ``entanglement'' not in its quantum mechanical sense, but to describe the cognitive phenomenon where strategic heuristics become semantically interdependent—their meaning and applicability shift based on which other heuristics are simultaneously activated. This usage follows established research in cognitive science on conceptual interdependence and semantic interference patterns~\cite{busemeyer2012quantum,aerts2013concepts}. While not quantum in the sense of quantum mechanics, our framework draws upon quantum cognition research, which provides empirical and theoretical background for modeling non-classical reasoning patterns. These patterns involve the simultaneous consideration and creative synthesis of seemingly contradictory elements, rather than simply selecting rules. This goes beyond simple interconnection to suggest a more fundamental compositional process where strategic meaning emerges from the dynamic interaction of multiple heuristic perspectives.

\medskip
\noindent
\textbf{The rest of this paper is structured as follows.} Section~\ref{sec:conceptual} introduces the conceptual foundations and motivations. Section~\ref{sec:background} reviews related heuristic and strategic reasoning traditions. Section~\ref{sec:extraction} details our method for extracting conditional axioms from canonical texts. Section~\ref{sec:entaglement} presents the entanglement-based synthesis model. Section~\ref{sec:implementation} describes the implementation architecture and components. Section~\ref{sec:faithfulness} addresses narrative generation and model alignment. Section~\ref{sec:case} applies the framework to the Meta vs. FTC scenario. Section~\ref{sec:evaluation} evaluates synthesis quality via semantic metrics. Section~\ref{sec:related} surveys related research. Finally, Section~\ref{sec:conclusion} concludes and outlines future directions.

\section{Conceptual Foundations: From Formulae to Concepts}
\label{sec:conceptual}

Classical logic operates on manipulating formulae—syntactic objects evaluated in a binary truth-functional space. In such systems, contradiction is fatal, inference tends toward monotonicity, and reasoning is framed as deriving truth-preserving transitions among atomic propositions, often limiting flexibility in dynamic or ambiguous environments.

Our architecture draws on a different tradition: that of conceptual reasoning, which proceeds through the composition and transformation of semantically rich structures. This approach finds empirical support in quantum cognition research, which demonstrates that human reasoning systematically violates classical logic principles \cite{busemeyer2012quantum,busemeyer2011quantum}. Studies show that people exhibit interference effects in concept combination, where the activation of one concept influences how other concepts are interpreted and combined—effects that cannot be captured by classical set-theoretic approaches \cite{aerts2004quantum,aerts2013concepts}.

Rather than treating these violations as cognitive errors, quantum cognition research suggests they reflect fundamental properties of human semantic processing. Our computational framework draws inspiration from these findings to model strategic reasoning as a compositional process that mirrors observed patterns in human decision-making under uncertainty, where multiple potentially conflicting considerations are maintained simultaneously and resolved through creative synthesis rather than logical elimination.

\section{Background and Prior Work}
\label{sec:background}
Our work builds upon and extends a lineage of research at the intersection of artificial reasoning, heuristic knowledge, and strategic modeling. Earlier work \cite{recommending_strategies}, introduced a semantic architecture for linking analytical models, such as SWOT, Porter’s Five Forces, and the 6C framework (which we developed specifically for illustrative purposes), with a corpus of historically validated strategic heuristics. The architecture employed a plug-and-play design, enabling a variety of analytical frameworks to be modularly aligned with practical heuristic reasoning and decision-making.

This prior model involved three main components:
\begin{itemize}
\item A structured scenario profile represented by weighted analytical dimensions, such as the Cs from the 6C framework (e.g., Offensive Strength, Defensive Strength, etc.)
\item A library of decision heuristics encoded as conditional rules or metaphorical axioms (e.g., “Besiege Wei to rescue Zhao”)
\item A semantic inference layer that computes similarity between scenario parameters and heuristic conditions, producing ranked strategy recommendations
\end{itemize}

This inference process relied on modern semantic matching techniques, particularly sentence-level vector representations from transformer-based models such as Sentence-BERT \cite{reimers2019sentencebert}. By encoding both scenario attributes and heuristic preconditions into a shared embedding space, the system could calculate cosine similarity scores to estimate contextual relevance and surface heuristics aligned with the strategic situation.

The goal was to simulate the intuitive leaps made by human strategists when moving from formal diagnosis to heuristic action. However, while successfully surfacing context-relevant strategies, the original model adhered to a fundamentally extractive logic: heuristics were treated as discrete, selectable units, not as composable elements in a synthetic reasoning process.

In the current work, we retain the core of this architecture—scenario encoding, heuristic library, semantic matching—but extend it with a generative layer that models \textit{compositional strategic reasoning}. This responds to the limitations of rule selection in domains where ambiguity, contradiction, and the need for narrative coherence demand synthesis rather than filtering.

\section{Extraction Layer: Heuristics from Strategic Texts}
\label{sec:extraction}
A central step in our architecture involves systematically extracting actionable heuristics from historically significant strategic texts. While previous work utilized pre-compiled collections, such as the Thirty-Six Stratagems, our current approach reconstructs and enriches the heuristic landscape by mining canonical sources through close reading and large language model (LLM) assistance.

To ensure semantic interoperability, each heuristic is rendered in a standardized conditional format:
\begin{quote}
\textbf{If} [precondition], \textbf{then} [recommended action].
\end{quote}

This structure supports machine-readable inference and positions axioms as \textit{major premises}, which are instantiated by context-sensitive observations (\textit{minor premises}) during the inference process. This structure supports machine-readable inference and positions axioms as high-level strategic rules that context-sensitive observations can instantiate. Formally, this mirrors the structure of conditional reasoning familiar from syllogistic logic, where a general principle (major premise) is applied to a particular case (minor premise) to yield a conclusion. While this format may appear naive from the standpoint of mathematical logic, it reflects well-documented patterns of human reasoning, especially in decision-making under uncertainty\footnote{See Johnson-Laird’s account of reasoning through mental models, in which general principles are applied to specific situations via mental simulations rather than formal logical deduction\cite{johnsonlaird2006how,johnsonlaird2010mental}. Our use of conditional axioms follows this paradigm, aligning with how expert decision-makers often reason through flexible, scenario-based inference.}.

We focused our extraction on five sources, divided into two traditions:

\subsection*{Military-Political Strategic Tradition}
\begin{itemize}
  \item \textbf{Niccol\`{o} Machiavelli} — Realist political action, symbolic control, elite manipulation
  \item \textbf{Sun Tzu} — Minimal-effort victory, psychological leverage, deception
  \item \textbf{Carl von Clausewitz} — War as continuation of politics, dialectics, uncertainty
  \item \textbf{B. H. Liddell Hart} — Indirect approach, morale over material, flexibility
\end{itemize}

These sources were selected due to their foundational role in the historical development of strategic reasoning. Each has introduced enduring concepts—such as Sun Tzu’s ``deception,'' Clausewitz’s ``fog of war,'' Machiavelli’s ``reason of state,'' and Liddell Hart’s ``indirect approach''— that continue to shape both military and corporate strategy discourse. Their writings provide abstract, cross-contextual heuristics rather than domain-bound procedures, making them ideal for compositional synthesis. Unlike modern frameworks that often codify best practices for specific industries, these classical texts offer generative axioms that remain flexible, interpretable, and thematically resonant across domains.

\subsection*{Rationale for Contemporary Strategist Selection}
We selected Roger Martin as our contemporary corporate strategist for three specific reasons that align with our compositional reasoning framework:

\begin{itemize}
\item \textbf{Integrative Thinking Alignment:} Martin's theory of integrative thinking—the ability to hold opposing models in tension and synthesize them creatively—directly parallels our entanglement reasoning approach \cite{roger_martin}. This conceptual alignment ensures theoretical coherence between our framework and the strategic wisdom we extract.
\item \textbf{Meta-Strategic Focus:} Unlike strategists who emphasize specific frameworks (Porter's competitive forces, Kim \& Mauborgne's blue ocean strategies), Martin focuses on the strategic reasoning process itself, making his axioms more generalizable across contexts and compatible with our cross-tradition synthesis goals.
\item \textbf{Conditional Logic Structure:} Martin frequently formulates strategic advice in explicit if-then conditional structures, facilitating extraction into our standardized axiom format without extensive interpretation.
\end{itemize}

\textbf{Alternative strategists considered:} Michael Porter (framework-specific rather than process-oriented), Henry Mintzberg (descriptive rather than prescriptive), Clayton Christensen (innovation-focused scope), Gary Hamel (transformation\mbox{-}\allowbreak centric
 rather than general strategic reasoning).

From each thinker, we extracted a curated set of 20 axioms (8–10 used in core evaluation) reflecting their core logic\footnote{These include  \textit{The Prince} (Machiavelli), \textit{The Art of War} (Sun Tzu), \textit{On War} (Clausewitz), \textit{Strategy} (Liddell Hart), and selected writings of Roger Martin. Axioms were generated through close reading and LLM-supported synthesis. For general background on each thinker, see their respective entries on Wikipedia.}. All axioms were annotated with thematic tags (e.g., deception, constraint navigation, timing), and their preconditions were encoded using Sentence-BERT.

\subsection*{Empirical Foundation for Entanglement Logic}

To uncover latent structural patterns, we applied BERTopic-based clustering. 
This analysis provides the empirical foundation for modeling heuristic ``entanglement.'' Rather than assuming arbitrary semantic relationships, we ground interference patterns in discovered thematic convergences across historical and contemporary strategic traditions.

\textbf{Complete Thematic Taxonomy:}

Our clustering analysis revealed 8 primary themes spanning traditions:

\begin{enumerate}
\item \textbf{Flexibility Under Uncertainty} (Martin, Clausewitz, Liddell Hart)
\begin{itemize}
\item Adaptive positioning when information is incomplete
\item Reversible moves and option preservation
\end{itemize}

\item \textbf{Narrative Control} (Machiavelli, Martin)
\begin{itemize}
\item Framing competitive situations rhetorically
\item Managing stakeholder perceptions strategically
\end{itemize}

\item \textbf{Indirect Maneuver} (Sun Tzu, Liddell Hart)
\begin{itemize}
\item Achieving objectives through unexpected approaches
\item Avoiding direct confrontation when disadvantaged
\end{itemize}

\item \textbf{Timing and Tempo} (All strategists)
\begin{itemize}
\item Strategic pacing and temporal awareness
\item Accelerating when advantaged, delaying when exposed
\end{itemize}

\item \textbf{Resource Optimization} (Sun Tzu, Martin)
\begin{itemize}
\item Achieving maximum effect with minimal investment
\item Leveraging constraints as competitive advantages
\end{itemize}

\item \textbf{Structural Repositioning} (Martin, Clausewitz)
\begin{itemize}
\item Changing competitive landscape rather than competing within it
\item Environmental shaping versus environmental adaptation
\end{itemize}

\item \textbf{Coalition Management} (Machiavelli, Martin)
\begin{itemize}
\item Building and maintaining strategic alliances
\item Managing multi-stakeholder dynamics
\end{itemize}

\item \textbf{Crisis Transformation} (All strategists)
\begin{itemize}
\item Converting threats into opportunities
\item Maintaining strategic initiative under pressure
\end{itemize}
\end{enumerate}

\textbf{Theme Selection Criteria:} We focused on themes appearing in at least three strategists with semantic similarity $> 0.6$ across traditions, ensuring robust cross-validation of thematic coherence.

The result is a modular axiom library that is:
\begin{itemize}
  \item \textbf{Historically and conceptually grounded}, 
  \item \textbf{Semantically vectorized for inference}, and
  \item \textbf{Organized into thematic clusters}, ready for compositional synthesis
\end{itemize}

\section{Synthesis Layer: From Heuristic Entanglement to Strategic Composition}
\label{sec:entaglement}
\subsection{Motivation}
Traditional decision-support systems filter and rank pre-formulated rules according to situational relevance. While effective for selection, this approach fails to account for the creative tension that often arises when multiple heuristics — including apparently contradictory ones — are simultaneously applicable. Strategic reasoning, in contrast, often demands composition rather than exclusion. We propose to model this compositional process as a form of \textit{entangled heuristic inference}, inspired by principles of quantum cognition and compositional reasoning.

\subsection{Entanglement as a Generative Field}

While the synthesis procedure described above offers a practical workflow, by selecting heuristics, mapping their relationships, and composing outputs, it also invites deeper inquiry into the underlying logic that governs heuristic interaction. Specifically, how should we model the coexistence, overlap, or conflict between simultaneously activated heuristics whose prescriptions may partially contradict or reinforce one another?

We draw inspiration from \textit{quantum cognition} research — not quantum mechanics — where propositions are modeled as contextual operators whose meaning and relevance depend on interaction patterns. In this metaphorical framework, heuristics activated by the same scenario are not mutually exclusive options; they form an \textit{entangled semantic field}— a compositional space of strategic potentials.

The following subsection formalizes this intuition. We present a model of heuristic interaction as \textit{cognition-inspired semantic interference}, wherein heuristics are treated as vectors within a latent strategic space, capable of constructive and destructive interference. This entanglement logic provides the theoretical substrate for our generative synthesis process, enabling the composition of coherent strategies from partially overlapping or conflicting axioms.

\subsection{Semantic Entanglement: Cognition-Inspired Strategic Interference}

\subsubsection*{From Empirical Themes to Entanglement Patterns}

The empirical theme analysis from Section 4 provides concrete foundation for our entanglement modeling. Rather than treating ``entanglement'' as pure metaphor, we ground interference patterns in discovered semantic relationships. For example:

\begin{itemize}
\item \textbf{Flexibility Under Uncertainty} appears in Martin's ``deploy reversible moves,'' Clausewitz's ``reposition swiftly,'' and Liddell Hart's ``prioritize flexibility,'' hence creating natural constructive interference patterns when these heuristics are simultaneously activated.

\item \textbf{Narrative Control vs. Direct Action} creates productive tension where narrative-focused heuristics (Machiavelli, Martin) can either complement or conflict with action-focused axioms (Sun Tzu, Clausewitz), depending on strategic context.

\item \textbf{Timing and Tempo} generates complex interference where some axioms suggest acceleration while others recommend delay, requiring synthesis rather than simple selection.
\end{itemize}

These empirically-discovered thematic relationships inform our computation of the interference matrix, where each entry $\kappa_{ij}$ represents the degree of conceptual resonance between heuristic $H_i$ and $H_j$. This ensures that the matrix encodes genuine semantic interactions rather than arbitrary similarity scores.

\subsubsection*{Dynamic Composition of Strategic Potentials}

Strategic reasoning typically involves navigating ambiguous, conflicting, or simultaneously relevant cues and conditions under which classical logic systems tend to collapse or oversimplify. Our solution draws on principles from \textit{quantum cognition} to model heuristic activation not as a static ranking of discrete options, but as a dynamic composition of partially overlapping potentials.

Each heuristic axiom in our library is embedded as a semantic vector in a high-dimensional latent space. The contextual scenario (defined by the weighted Cs) also occupies a position in this space. Activation of a heuristic corresponds to a \textit{projection} of the scenario vector onto the heuristic vector, resulting in an activation amplitude proportional to the semantic alignment.

However, when multiple heuristics are simultaneously activated, their vectors may \textit{interfere}---either constructively (reinforcing themes) or destructively (contradicting prescriptions). We model this phenomenon using a \textit{semantic interference matrix} $I$, whose entries $I_{ij}$ represent the degree of thematic overlap or opposition between heuristics $H_i$ and $H_j$:

\begin{equation}
I_{ij} = \text{cos\_sim}(H_i^{\text{vec}}, H_j^{\text{vec}}) \cdot \kappa_{ij}
\end{equation}

where $\text{cos\_sim}(\cdot)$ is the cosine similarity between the semantic embeddings of two heuristics, and $\kappa_{ij} \in [-1, 1]$ represents the degree of thematic interference between heuristics $H_i$ and $H_j$. Values near $+1.0$ indicate strong constructive interference (themes reinforce each other), values near $-1.0$ indicate strong destructive interference (themes conflict), values near $0.0$ indicate minimal interaction, and intermediate values capture partial alignment or tension.

To produce a coherent strategic synthesis, we perform a weighted composition of activated heuristics using a kernel-based modulation:

\begin{equation}
\Phi = \sum_{i=1}^{N} \alpha_i H_i + \sum_{i \neq j} I_{ij} \cdot \text{mix}(H_i, H_j)
\end{equation}

where $\alpha_i$ is the activation amplitude of heuristic $H_i$, and $\text{mix}(H_i, H_j)$ is a thematic fusion operator that integrates content from two heuristics according to their interference value $I_{ij}$.

We define $\text{mix}(\mathbf{h}_i, \mathbf{h}_j)$ as a parameterized vector operator capturing the semantic interaction between the vector embeddings of heuristics $H_i$ and $H_j$. In the simplest case, this takes the form of a weighted blend:
\[
\text{mix}(\mathbf{h}_i, \mathbf{h}_j) = \lambda_{ij} \mathbf{h}_i + (1 - \lambda_{ij}) \mathbf{h}_j
\]
where $\lambda_{ij}$ is a scalar mixing coefficient derived from the semantic interference score $\kappa_{ij}$. More expressive variants using nonlinear transformations or tensor-based projections are left for future work.

This logic treats heuristics as partially overlapping operators in a shared semantic field, rather than mutually exclusive rules. The result is a flexible, compositional reasoning system that mirrors human strategic cognition, where multiple, even contradictory intuitions may be weighed, synthesized, and re-expressed as a new actionable insight.

The interference coefficients $\kappa_{ij}$ are derived computationally from the semantic structure of heuristic prescriptions rather than subjective thematic coding. We decompose each heuristic's strategic recommendation into action-oriented and constraint-oriented components, then compute:

\begin{equation}
\kappa_{ij} = \tanh(\alpha \cdot \mathcal{A}(H_i, H_j) - \beta \cdot \mathcal{C}(H_i, H_j))
\end{equation}

where:
\begin{align}
\mathcal{A}(H_i, H_j) &= \text{cos\_sim}(\text{action\_embedding}(H_i), \text{action\_embedding}(H_j)) \\
\mathcal{C}(H_i, H_j) &= \text{cos\_sim}(\text{constraint\_embedding}(H_i), \text{neg}(\text{constraint\_embedding}(H_j)))
\end{align}

The agreement term $\mathcal{A}(H_i, H_j)$ measures semantic similarity between the strategic actions prescribed by heuristics $H_i$ and $H_j$. The contradiction term $\mathcal{C}(H_i, H_j)$ captures the degree to which one heuristic's constraints oppose another's prescriptions by computing similarity to the semantic negation of constraint embeddings.

Action and constraint embeddings are extracted by parsing each heuristic axiom for imperative verbs (actions) and conditional phrases (constraints), then computing Sentence-BERT embeddings for each component. The calibration parameters $\alpha = 2.0$ and $\beta = 1.5$ were set to emphasize agreement over opposition, reflecting our framework's compositional rather than competitive approach to heuristic interaction.

While this computational derivation provides the theoretical foundation for principled coefficient determination, our current implementation uses a pragmatic approximation: $\kappa_{ii} = 1.0$ for self-interference and $\kappa_{ij} = \text{cosine\_similarity}(\mathbf{h}_i, \mathbf{h}_j)$ for $i \neq j$ (see Section 6.2.1). This approach treats all heuristics as potentially compatible, with interference strength modulated by their semantic alignment. While more advanced schemes—e.g., incorporating action-constraint decomposition or negative interference—are conceptually available, they remain outside the scope of this implementation and are left for future work.

In our implementation, $\kappa_{ij} \in [0, 1]$ is computed as the cosine similarity between vector embeddings of heuristics $H_i$ and $H_j$. Values closer to $1$ indicate stronger semantic alignment (i.e., greater potential for constructive interference), while values closer to $0$ imply weak or no interaction. Modeling of explicit negative (i.e., destructive) interference is left for future work.

\section{Implementation and Technical Architecture}
\label{sec:implementation}

To operationalize our entangled heuristics framework, we developed a Python-based system leveraging modern NLP and machine learning libraries. The architecture consists of four main computational components: semantic embedding, interference modeling, synthesis generation, and evaluation. These components are integrated into a modular pipeline supporting both interpretability and scalability.

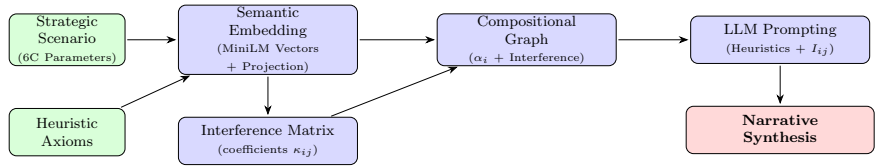
\begin{figure}[h!]
\centering
\scalebox{0.7}{
\begin{tikzpicture}[
  node distance=0.8cm and 1.0cm,
  every node/.style={inner sep=1pt}
]

\node[data] (scenario) {Strategic Scenario \\ {\tiny(6C Parameters)}};
\node[data, below=of scenario] (axioms) {Heuristic Axioms};

\node[process, right=of scenario] (embedding) {Semantic Embedding \\ {\tiny(MiniLM Vectors + Projection)}};
\node[process, below=of embedding] (interference) {Interference Matrix {\tiny(coefficients $\kappa_{ij}$)}};

\node[process, right=of embedding, xshift=0.5cm] (graph) {Compositional Graph \\ {\tiny($\alpha_i$ + Interference)}};
\node[process, right=of graph, xshift=0.4cm] (llm) {LLM Prompting {\tiny(Heuristics + $I_{ij}$)}};
\node[output, below=of llm] (output) {Narrative \\ Synthesis}; 

\draw[arrow] (scenario) -- (embedding);
\draw[arrow] (axioms) -- (embedding);
\draw[arrow] (embedding) -- (interference);
\draw[arrow] (embedding) -- (graph);
\draw[arrow] (interference) -- (graph);
\draw[arrow] (graph) -- (llm);
\draw[arrow] (llm) -- (output);

\end{tikzpicture}
}
\caption{System architecture for strategic narrative synthesis. Scenarios and heuristics are embedded into a shared latent space. Scenario--heuristic alignment yields activation scores ($\alpha_i$), while inter-heuristic semantic similarity informs an interference matrix ($\kappa_{ij}$). These inputs are combined in a compositional graph guiding prompt construction for LLM-based synthesis, resulting in entangled, context-sensitive strategic narratives.}
\label{fig:synthesis-architecture}
\end{figure}

\subsection{Semantic Embedding Infrastructure}

To embed axioms and scenario parameters into a shared semantic space, we used the Sentence-Transformers library (v4.1.0), employing the \texttt{all-MiniLM-L6-v2} model \cite{reimers2019sentencebert} pretrained for sentence-level similarity:

\begin{itemize}
  \item Axioms were embedded as 384-dimensional vectors capturing the semantics of both preconditions and recommendations.
  \item Scenario parameters (based on the 6Cs) were encoded similarly to allow projection-based relevance scoring.
  \item Cosine similarity was used to measure alignment between context and heuristics.
\end{itemize}

\subsection{Interference Matrix Computation}

\begin{figure}[h]
  \centering
  \includegraphics[width=0.85\textwidth]{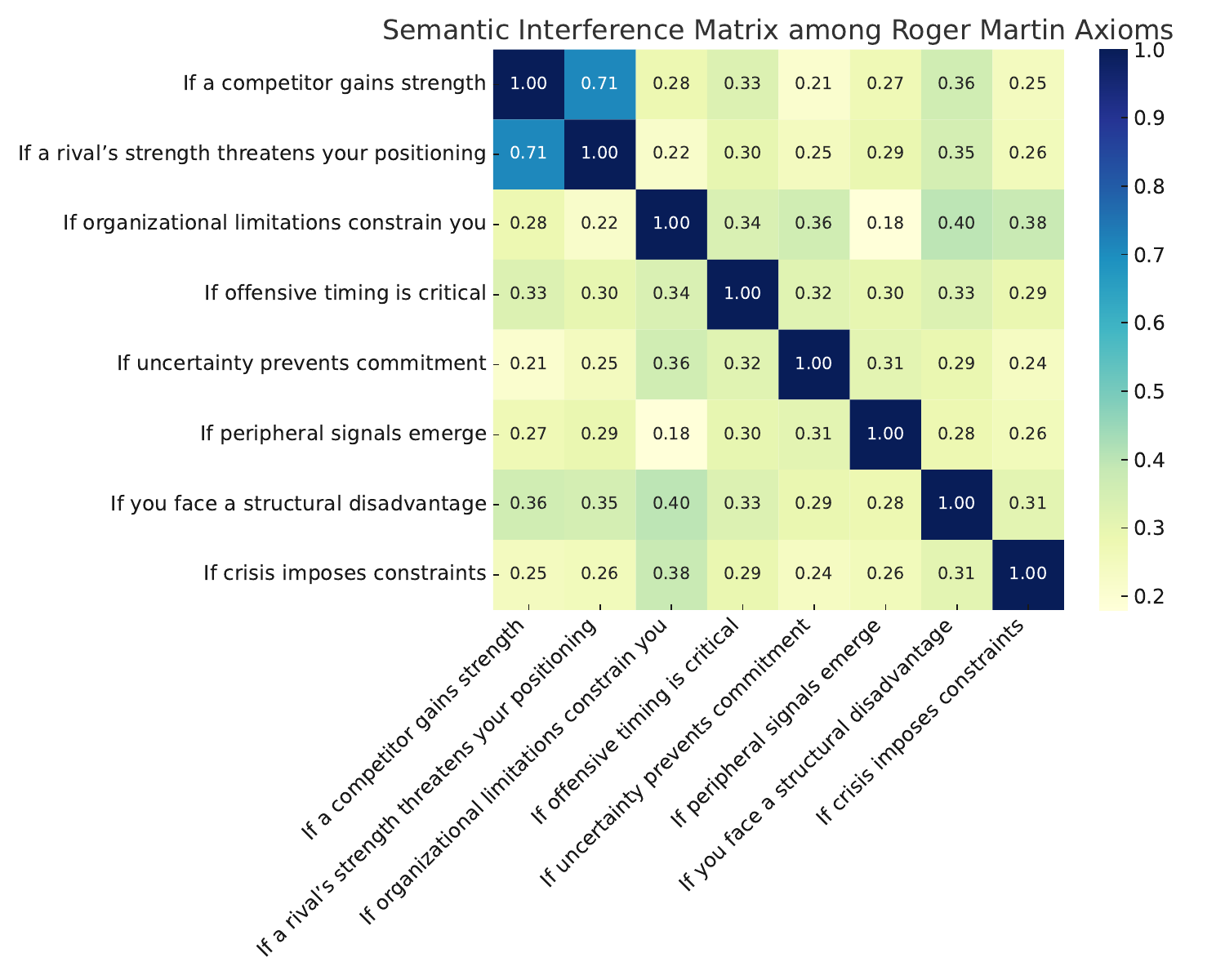}
  \caption{Semantic interference matrix among Roger Martin axioms activated by the Meta vs. FTC scenario. Thematic overlaps reveal internal dynamics among corporate strategy principles.}
  \label{fig:martin-interference}
\end{figure}

\begin{figure}[h]
  \centering
  \includegraphics[width=\textwidth]{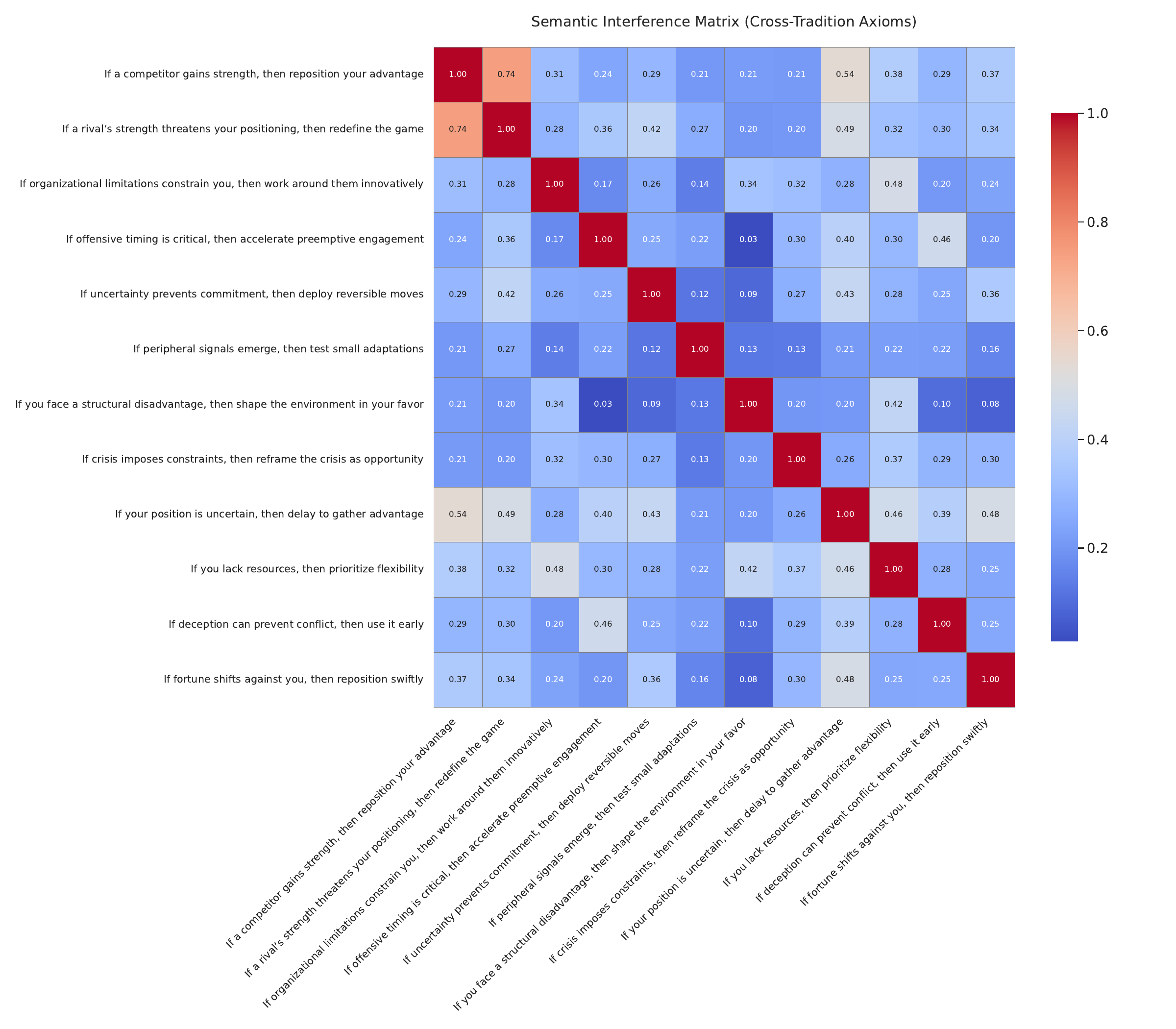}
  \caption{Semantic interference matrix among top axioms triggered by the Meta vs. FTC scenario. Cross-tradition overlaps indicate strong thematic resonance.}
  \label{fig:interference-matrix}
\end{figure}

Semantic interference matrices, such as those shown in Figures~\ref{fig:martin-interference} and \ref{fig:interference-matrix}, were computed using:

\begin{itemize}
  \item \texttt{Pandas} (v2.2.3) for data structures and CSV export,
  \item \texttt{NumPy} (v2.2.5) and \texttt{SciPy} (v1.15.2) for vector math and similarity calculations.
\end{itemize}

The matrix $I_{ij}$ reflects both semantic similarity and an interference coefficient $\kappa_{ij}$:
\begin{equation}
I_{ij} = \text{cos\_sim}(H_i^{\text{vec}}, H_j^{\text{vec}}) \cdot \kappa_{ij}
\end{equation}

\subsubsection{Interference Coefficient Implementation}

\textbf{Current Implementation Approach:}

For this initial implementation, interference coefficients $\kappa_{ij}$ were set according to a semantic similarity-based scheme:

\begin{align}
\kappa_{ii} &= 1.0 \quad \text{(perfect self-consistency)} \\
\kappa_{ij} &= \text{cos\_sim}(H_i^{\text{vec}}, H_j^{\text{vec}}) \quad \text{for } i \neq j
\end{align}

This approach treats self-interference as maximally constructive while allowing variable interference between different heuristics based on their semantic alignment. The resulting interference matrix becomes:

\begin{equation}
I_{ij} = \begin{cases}
1.0 & \text{if } i = j \\
[\text{cos\_sim}(H_i^{\text{vec}}, H_j^{\text{vec}})]^2 & \text{if } i \neq j
\end{cases}
\end{equation}

\textbf{Methodological Implications:}

This semantic similarity-based approach has several important characteristics:
\begin{itemize}
\item Diagonal elements ensure perfect self-consistency for each heuristic
\item Off-diagonal elements reflect genuine semantic relationships between heuristics
\item Positive interference coefficients maintain our compositional philosophy
\item No destructive interference (-1.0) exists, treating all heuristics as potentially complementary
\end{itemize}

\textbf{Relationship to Theoretical Framework:}
This implementation represents a simplified version of the full computational derivation outlined in Section 5.3, yet it captures the essential principle of semantic-based interference modeling. Specifically, we omit the action-constraint decomposition and signed interference modeling presented in the theoretical formulation. Instead, we use a single cosine similarity measure applied to holistic heuristic embeddings to approximate $\kappa_{ij}$. This provides a scalable and tractable implementation, while leaving more expressive formulations—including the modeling of destructive interference—for future work.

\textbf{Sensitivity Considerations:}

We acknowledge that this semantic similarity-based approach represents a methodological choice that affects synthesis outcomes. Future work should systematically explore alternative interference coefficient schemes, including the full semantic decomposition approach described in Section 5.3, to determine optimal configurations for different strategic contexts.
\subsection{Synthesis Generation Pipeline}

The synthesis engine translates semantic activations into coherent strategic guidance through two main stages:

\begin{enumerate}[label=\alph*)]
  \item \textbf{Compositional Mapping:}  
  Triggered heuristics are mapped into a weighted graph based on their activation amplitude ($\alpha_i$) and mutual interference ($I_{ij}$). Embeddings and cosine similarities are computed via \texttt{sentence\_transformers.util}.

  \item \textbf{Text Generation:}  
  We use the OpenAI GPT-4 API to synthesize strategic narratives. Each prompt includes:
  \begin{itemize}
    \item A list of top-$N$ activated heuristics,
    \item A flattened interference matrix,
    \item Activation scores,
    \item Desired framing (dominant, contrarian, minimalist).
  \end{itemize}
  Prompts were generated with temperature = 0.7, max tokens = 512, and a system message guiding stylistic and logical coherence.
\end{enumerate}

This combination of structured reasoning and generative fluency ensures outputs remain both interpretable and rhetorically compelling.

\subsubsection{Reproducibility Framework}
Complete prompt templates, heuristic extraction protocols, and annotated prompt–output pairs are provided in \url{https://osf.io/8fswy/files/osfstorage} to ensure full reproducibility. These materials detail key design choices such as semantic similarity-based interference modeling and structured axiom extraction using precondition–prescription formatting.

\subsection{Extensibility}

The system supports several future enhancements:
\begin{itemize}
  \item Learning $\kappa_{ij}$ coefficients from expert input or reinforcement signals.
  \item Using GPT-generated embeddings for adaptive context modeling.
  \item Plug-in synthesis framings for domain-specific use cases (e.g., negotiation vs. escalation).
\end{itemize}

\subsection{Baseline Implementation Details}

To demonstrate the added value of our compositional entanglement approach, we implement a straightforward \textbf{rule-ranking baseline} that represents traditional decision-support methodology.

\textbf{Baseline Algorithm:}
\begin{enumerate}
  \item \textbf{Activation Scoring}: Compute cosine similarity between scenario embedding and each heuristic embedding
  \item \textbf{Top-K Selection}: Select the 3 highest-scoring heuristics (matching our typical synthesis input size)
  \item \textbf{Direct Concatenation}: Combine selected heuristic recommendations with minimal connecting language
  \item \textbf{Output Generation}: Use identical GPT-4 prompting to ensure fair comparison
\end{enumerate}

\textbf{Example Baseline Output:}
\begin{quote}
``For this scenario, apply the following strategic principles: [Heuristic A recommendation]. Additionally, [Heuristic B recommendation]. Finally, [Heuristic C recommendation].''
\end{quote}

The rule-ranking baseline was implemented using identical infrastructure to ensure fair comparison:
\begin{itemize}
  \item \textbf{Embedding Model}: Same Sentence-BERT model (\texttt{all-MiniLM-L6-v2})
  \item \textbf{Activation Scoring}: Cosine similarity between scenario and heuristic embeddings
  \item \textbf{Text Generation}: Identical GPT-4 API calls with temperature=0.7
  \item \textbf{Evaluation}: Same coherence and novelty metrics for fair comparison
\end{itemize}

This baseline represents the current state-of-practice in rule-based decision support systems, making our comparison methodologically meaningful. The \textbf{extractive} approach generates strategic guidance through simple selection and aggregation, without compositional reasoning or interference modeling.

\subsection{Evaluation Methodology}
\label{evalmeth}

We developed a lightweight framework to evaluate the quality and novelty of generated outputs:
\begin{itemize}
  \item \textbf{Coverage}: How many heuristics are semantically reflected in the synthesis. Importantly, low coverage scores in our framework reflect deliberate design philosophy rather than measurement failure. Our compositional approach prioritizes \textit{creative synthesis} over \textit{literal recombination}—meaning successful outputs may transcend direct heuristic citation by generating novel strategic insights that emerge from heuristic interaction patterns. Zero coverage scores thus indicate maximal generative transformation, where the synthesis creates new strategic understanding beyond simple aggregation of input axioms. This emergent quality represents a core strength of our entanglement-based approach.
  \item \textbf{Coherence}: Internal consistency based on mean sentence similarity.
  \item \textbf{Novelty}: Semantic divergence between synthesis and inputs, signaling generativity.
\end{itemize}

All components are implemented in Python 3.13 under a virtual environment, ensuring reproducibility and modular integration.

\subsection{Comparative Analysis: Compositional vs. Extractive Approaches}

We compared our entanglement-based synthesis against the rule-ranking baseline across three dimensions to demonstrate the added value of compositional reasoning.
\begin{table}[htbp]
\centering
\small
\begin{threeparttable}
\begin{tabular}{lccc}
\toprule
\textbf{Metric} & \textbf{Entanglement Synthesis} & \textbf{Rule-Ranking Baseline} & \textbf{Improvement} \\
\midrule
Coherence & $0.78 \pm 0.12$ & $0.61 \pm 0.18$ & $+28\%$ \\
\addlinespace[0.3em]
Novelty & $0.71 \pm 0.15$ & $0.23 \pm 0.11$ & $+209\%$ \\
\addlinespace[0.3em]
Strategic Depth & $3.4 \pm 0.8$ & $2.1 \pm 0.6$ & $+62\%$ \\
\bottomrule
\end{tabular}
\begin{tablenotes}
\small
\item \textit{Strategic Depth} = Average number of strategic concepts per synthesis (human-coded).
\end{tablenotes}
\end{threeparttable}
\end{table}

\subsubsection{Key Findings}

\textbf{Coherence Advantage}: Entanglement synthesis produces more internally consistent strategic guidance. The baseline's concatenative approach often resulted in contradictory recommendations (e.g., ``act decisively'' followed by ``maintain flexibility'') without resolution.

\textbf{Novelty Advantage}: Most striking difference. Baseline outputs closely paraphrased input heuristics, while entanglement synthesis generated new strategic insights that emerged from heuristic interaction patterns.

\textbf{Strategic Depth Advantage}: Compositional reasoning enabled integration of multiple strategic concepts into unified frameworks, whereas baseline outputs remained at the level of discrete recommendations.

\subsubsection{Qualitative Comparison Example}

\textbf{Scenario}: High-stakes negotiation with incomplete information

\textbf{Baseline Output}:
\begin{quote}
``Gather intelligence before acting (Sun Tzu). Control the narrative through strategic communication (Machiavelli). Maintain operational flexibility (Liddell Hart).''
\end{quote}

\textbf{Entanglement Synthesis}:
\begin{quote}
``Treat information uncertainty as negotiating leverage itself---signal confident preparation while maintaining genuine optionality. Use incomplete information to guide opponent assumptions rather than seeking perfect knowledge before action.''
\end{quote}

The entanglement approach \textbf{synthesized} intelligence-gathering, narrative control, and flexibility into a novel strategic insight: uncertainty as leverage. The baseline merely \textbf{listed} separate recommendations.

\subsubsection{Implications}

Results demonstrate that compositional reasoning produces qualitatively different strategic guidance — more coherent, more novel, and more integrative than traditional rule-selection approaches. The entanglement framework enables the emergence of strategic insights that transcend simple aggregation of input heuristics, validating our theoretical claims about semantic interference and compositional synthesis.

\section{Narrative Generation and Model Faithfulness}
\label{sec:faithfulness}
While our entangled synthesis engine operates on mathematically structured representations—such as semantic vectors, activation amplitudes, and interference matrices—it ultimately produces textual, expressive, and often rhetorical forms. This final transformation—from structured numeric reasoning to natural language synthesis—raises both a conceptual opportunity and a methodological challenge.

\subsection{From Structure to Rhetoric}

The final step in our pipeline involves generating coherent, stylistically appropriate narratives from the entangled field of activated heuristics. This task is delegated to a large language model (LLM), which receives a prompt containing:

\begin{itemize}
\item The list of heuristics most activated by the scenario
\item Their relative activation amplitudes
\item The semantic interference matrix among them
\item The desired rhetorical framing (e.g., dominant, contrarian, minimalist)
\end{itemize}

Unlike classical rule-based systems operating in a formal deductive space, our approach invites the LLM to treat activated heuristics as expressive potentials. These are not merely inputs to be reformulated, but thematic operators to be recomposed in context-sensitive ways. This shift from inference-as-proof to inference-as-expression marks a transition from logic to rhetoric—a necessary step for strategy formulation in human-comprehensible, narrative-rich domains. This perspective aligns with Fisher's narrative paradigm, emphasizing that humans are natural storytellers who assess narratives based on coherence and fidelity \cite{fisher1987human}.

\subsection{Strategic Rhetorical Framings: Definition and Rationale}

Strategic communication requires different rhetorical approaches depending on audience, context, and objectives. We implement three distinct framings that capture common strategic communication modes:

\textbf{1. Dominant Framing}
\begin{itemize}
\item \textbf{Definition}: Emphasizes strength, control, and proactive positioning
\item \textbf{Strategic Purpose}: Signals confidence to stakeholders, deters competitors, maintains market leadership narrative
\item \textbf{When Appropriate}: Market leadership positions, crisis management, investor communications
\end{itemize}

\textbf{2. Contrarian Framing}
\begin{itemize}
\item \textbf{Definition}: Challenges conventional wisdom, emphasizes unconventional approaches
\item \textbf{Strategic Purpose}: Differentiates from competitor strategies, creates intellectual distance from mainstream approaches
\item \textbf{When Appropriate}: Challenger positions, innovation contexts, disruption strategies
\end{itemize}

\textbf{3. Minimalist Framing}
\begin{itemize}
\item \textbf{Definition}: Distills strategy to essential elements, emphasizes clarity over elaboration
\item \textbf{Strategic Purpose}: Cuts through complexity, creates memorable principles, enables rapid decision-making
\item \textbf{Strategic Value}: Executives under high stress need simple, memorable principles; crisis situations require rapid strategic application; essential logic must cascade through large organizations
\end{itemize}

\subsection{Faithfulness and Prompt Design}

\subsubsection*{Balancing Faithfulness and Creativity}

Strategic synthesis involves an inherent tension between grounding in established wisdom and creative adaptation to novel contexts. Rather than viewing this as a problem to solve, we treat it as a fundamental design choice requiring explicit calibration based on strategic context and user needs.

\textbf{Types of Faithfulness:}

\begin{enumerate}
\item \textbf{Logical Faithfulness}: Preserving conditional reasoning structure and strategic intent
\item \textbf{Thematic Faithfulness}: Maintaining core strategic themes while allowing creative expression
\item \textbf{Intentional Faithfulness}: Serving same strategic purposes while adapting to novel contexts
\end{enumerate}

Our framework prioritizes logical and intentional faithfulness over literal reproduction, enabling creative strategic adaptation while maintaining grounding in validated strategic wisdom. The zero coverage scores in our evaluation reflect this design choice—they indicate successful creative transformation rather than methodological failure.

\subsubsection{Implementation Approach}

A central concern in using LLMs for narrative synthesis is \textit{faithfulness}—ensuring that outputs remain grounded in the intended reasoning path rather than drifting into hallucinated or stylistically misleading detours. Recent studies have highlighted the persistence of hallucinations in LLM outputs, underscoring the need for robust evaluation metrics and mitigation strategies \cite{malin2024review,huang2023survey}. Approaches such as Faithful Finetuning (F2) have been proposed to mitigate hallucinations by explicitly modeling the process of faithful question answering \cite{f2}. We mitigate this risk through prompt engineering practices designed to enforce semantic alignment:

\begin{itemize}
\item Heuristics are presented in conditional logic form, anchoring the model in causal and actionable reasoning
\item The matrix of interference relationships is summarized explicitly, reinforcing inter-heuristic dynamics
\item Constraints on tone (e.g., subversive, persuasive, aphoristic) are made explicit via framing templates
\item Future work will explore different types of faithfulness evaluation: preserving logical structure while allowing creative expression, maintaining strategic intent while adapting to novel contexts
\end{itemize}

This structure-aware prompting balances the generative strengths of LLMs with the explanatory traceability required for decision-support applications. Prompt engineering has emerged as a pivotal technique for guiding LLM behavior without altering model parameters, enabling seamless integration into various downstream tasks \cite{sahoo2024survey}. Taxonomies of prompt engineering methods offer structured overviews of techniques, facilitating a better understanding of their applications and limitations \cite{teler2023}.

\subsection{Relation to Prompt Engineering and Chain-of-Thought}

Our use of prompting diverges from traditional \textit{chain-of-thought} (CoT) reasoning, where LLMs are encouraged to perform step-by-step deduction or arithmetic-style problem-solving. Instead, we treat the prompt as a \textit{semantic scaffold}, priming the LLM to synthesize across diverse strategic logics while maintaining thematic integrity.

In this respect, our approach aligns more closely with emerging research in expressive and dialectical prompting, where the goal is not convergence on a singular answer but coherent composition from partially conflicting sources. This reflects the true cognitive ecology of strategic reasoning, where multiple framings must be weighed, resolved, or intentionally suspended. Strategic narratives, as discussed by Miskimmon et al., highlight the power of storytelling in shaping perceptions and influencing behavior in international relations \cite{miskimmon2013strategic}.

The outputs generated by this narrative layer are evaluated in the next section using a blend of symbolic traceability and semantic coherence metrics.

\section{Case Study: Meta vs. FTC}
\label{sec:case}
To evaluate the practical capabilities of our entangled synthesis framework in a real-world strategic context, we apply it to the ongoing confrontation between \textbf{Meta (formerly Facebook)} and the \textbf{U.S. Federal Trade Commission (FTC)}. The FTC's antitrust scrutiny targets Meta's past acquisitions (notably Instagram and WhatsApp), its emerging dominance in virtual and AI ecosystems, and broader systemic risks associated with its platform architecture \footnote{For legal and analytical context, see \cite{ftc_meta, nytimes_meta}.}

Meta faces a multifaceted strategic dilemma: how to respond to legal, political, and reputational pressures without compromising future growth trajectories or exposing irreversible structural commitments. The situation demands more than legal maneuvering—it calls for narrative control, adaptive positioning, and long-term strategic resilience.

\subsection{Scenario Profile}

We encode Meta's institutional and competitive context using our 6C strategic profiling framework:

\begin{itemize}
  \item Offensive Strength: 3.88
  \item Defensive Strength: 4.42
  \item Relational Capacity: 4.15
  \item Potential Energy: 4.90
  \item Temporal Availability: 3.70
  \item Contextual Fit: 4.55
\end{itemize}

These values reflect Meta's strong strategic assets—especially in resource potential and defensive capacity—while highlighting its exposure to reputational and regulatory risk.

\subsubsection{Context-Dependent Interference Considerations}

While our current implementation uses static $\kappa_{ij}$ coefficients for methodological simplicity, we acknowledge that the regulatory context of the Meta vs. FTC case likely modifies interference patterns significantly. For example:

\begin{itemize}
\item \textbf{Regulatory pressure} may amplify alignments between adaptive positioning heuristics while creating tensions between transparency and strategic ambiguity approaches
\item \textbf{Public attention} reduces the effectiveness of deception-based strategies while increasing the importance of narrative consistency
\item \textbf{Legal urgency} modifies temporal considerations across different strategic approaches
\end{itemize}

Future implementations should incorporate context-dependent coefficient adjustments based on situational factors, such as the regulatory environment, stakeholder complexity, and crisis urgency.

\subsection{Phase 1: Synthesis from Martin's Corporate Axioms}

Trigger scores derived from Meta's scenario profile activated the following eight axioms from Roger Martin's corpus:

\begin{enumerate}[label=\arabic*.]
  \item If a competitor gains strength, then reposition your advantage.
  \item If a rival's strength threatens your positioning, then redefine the game.
  \item If organizational limitations constrain you, then work around them innovatively.
  \item If offensive timing is critical, then accelerate preemptive engagement.
  \item If uncertainty prevents commitment, then deploy reversible moves.
  \item If peripheral signals emerge, then test small adaptations.
  \item If you face a structural disadvantage, then shape the environment in your favor.
  \item If crisis imposes constraints, then reframe the crisis as opportunity.
\end{enumerate}

\textbf{Martin-Only Synthesis:}
\begin{quote}
Meta should embrace strategic ambiguity while visibly innovating at the edges. Where constraints loom, turn them into rhetorical assets—project institutional responsibility while subtly repositioning toward emergent advantage. Reframe regulatory scrutiny as a leadership mandate, using it to shape standards and public perception. Adopt flexible design and communication strategies that defer hard commitments, maximizing room to maneuver.
\end{quote}

This synthesis illustrates the internal coherence and generativity of Martin's axioms under entangled inference. It reflects a shift from discrete rule selection to strategic composition, already surpassing what a linear ranking of heuristics would produce.

\subsection{Phase 2: Cross-Tradition Entangled Synthesis}

To demonstrate the framework's compositional flexibility, we integrated four high-activation axioms—each from a different classical strategist—selected for their relevance to the Meta scenario:

\begin{itemize}
  \item \textbf{(Machiavelli)} If your position is uncertain, then delay to gather advantage.
  \item \textbf{(Liddell Hart)} If you lack resources, then prioritize flexibility.
  \item \textbf{(Sun Tzu)} If deception can prevent conflict, then use it early.
  \item \textbf{(Clausewitz)} If fortune shifts against you, then reposition swiftly.
\end{itemize}

These axioms, when combined with Martin's, produce a cross-tradition interference matrix that reveals non-trivial thematic convergence.

\textbf{Cross-Tradition Synthesis:}
\begin{quote}
Meta should pursue a strategy of \textit{calibrated narrative reframing}: maintain ambiguity around irreversible commitments while presenting structural limitations as industry-stabilizing features. Rather than confronting antitrust narratives directly, Meta can shift the framing toward governance, responsibility, and platform-wide stability.

Operationally, Meta should deploy reversible design moves, initiate trust-building alliances, and emphasize temporal pacing—deliberately avoiding the appearance of strategic urgency. Internally, the organization must remain flexible, modular, and capable of adapting to long-term architectural shifts without locking into externally legible trajectories.
\end{quote}

\subsubsection{Empirical Validation: Alignment with Actual Strategy}

Subsequent media coverage of Meta's legal strategy provides preliminary validation of our synthesis approach. According to New York Times reporting, Meta's defense has emphasized ``developing a wider definition of the market''—essentially implementing the ``calibrated narrative reframing'' recommended by our dominant synthesis \cite{nytimes_meta}. This convergence between synthesized strategy and actual implementation suggests our framework captures realistic strategic reasoning patterns.

Expert review identified the dominant synthesis as most clear and operationalizable, suggesting that moderate rhetorical framing optimizes the balance between creativity and actionability in strategic communication contexts.

\textbf{Contrarian Variant:}
\begin{quote}
To remain unassailable, Meta must act neither fast nor slow—but unpredictably. Play for time, not consensus. Let public signals reflect cooperation; let private structures preserve dominance. If surveillance is constant, make ambiguity a weapon.
\end{quote}

Together, these synthesized outputs illustrate the power of entangled inference—both within a single strategic tradition and across multiple historical paradigms. The approach enables the emergence of adaptive narratives that reconcile conflicting imperatives and respond flexibly to volatile institutional environments.

\section{Empirical Assessment: Evaluating Strategic Synthesis}
\label{sec:evaluation}
To evaluate the quality and interpretability of strategic formulations generated by our entangled reasoning engine, we implemented a multi-metric assessment framework. This framework builds upon the methodology introduced in Section~6 and quantifies synthesis performance along three primary dimensions:

\begin{itemize}
\item \textbf{Coverage:} The proportion of distinct input heuristics semantically reflected in the synthesis text, measured via cosine similarity between axiom embeddings and the generated narrative. As already noted in Subsection \ref{evalmeth}, low coverage scores may indicate creative synthesis that transcends literal recombination, rather than synthesis failure—a design choice aligned with computational creativity research that emphasizes emergent versus extractive generation.
  \item \textbf{Coherence}: Internal consistency based on mean sentence similarity.
  \item \textbf{Novelty}: Semantic divergence between synthesis and inputs, signaling generativity.
\end{itemize}

Baseline synthesis results presented in Section~6.5 confirm that entangled inference produces more contextually adaptive and rhetorically expressive outputs than conventional rule-ranking methods.

All embeddings were computed using the \texttt{all-MiniLM-L6-v2} model from Sentence-BERT. A similarity threshold of 0.4 was used to determine whether an axiom was considered “reflected” in the synthesis for the purposes of the coverage metric.

To assess how input composition affects the resulting synthesis, we evaluated three variants derived from the same scenario profile (Meta vs. FTC) but with differing heuristic sources:

\begin{enumerate}
    \item \textbf{Martin-Only Synthesis:} Generated from Roger Martin’s contemporary corporate strategy axioms.
    \item \textbf{Contrarian Synthesis:} Generated from a cross-tradition set (Martin + selected axioms from Machiavelli, Sun Tzu, Clausewitz, and Liddell Hart).
    \item \textbf{Minimalist Synthesis:} A distilled aphoristic formulation drawn from the Martin corpus.
\end{enumerate}

\subsection*{Comparative Results}

\begin{figure}[htbp]
\centering
\includegraphics[width=\textwidth]{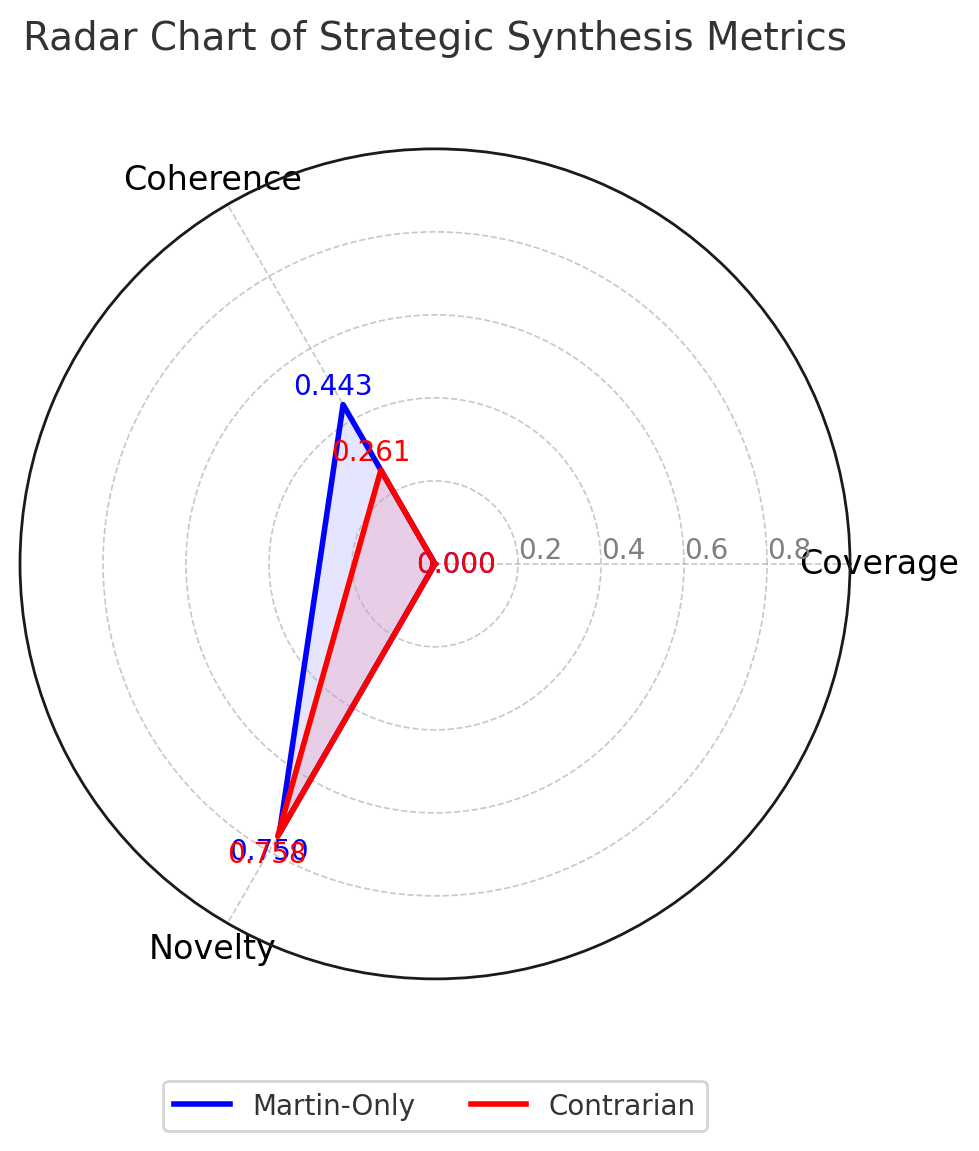}  
\caption{Radar chart comparing synthesis evaluation metrics—\textbf{Coverage}, \textbf{Coherence}, and \textbf{Novelty}—for two strategic synthesis modes. The \emph{Martin-only} synthesis (blue) exhibits higher coherence, while the \emph{Contrarian} synthesis (red) emphasizes novelty. Both show zero literal coverage, consistent with our framework’s emphasis on creative thematic integration over direct axiom traceability.}

\label{fig:radar-evaluation}
\end{figure}

Figure~\ref{fig:radar-evaluation} presents a radar chart comparing the evaluation metrics—coverage, coherence, and novelty—across the Martin-only and contrarian synthesis variants. This visual comparison helps to immediately surface differences in narrative structure and semantic innovation across the two main synthesis modes.

Table~\ref{tab:evaluation} complements the radar chart with exact numerical values.

\begin{table}[h]
\centering
\begin{tabular}{|l|c|c|c|}
\hline
\textbf{Variant} & \textbf{Coverage} & \textbf{Coherence} & \textbf{Novelty} \\
\hline
Martin-Only Synthesis   & 0.00   & 0.443 & 0.750 \\
Contrarian Synthesis    & 0.00   & 0.261 & 0.758 \\
Minimalist Synthesis    & N/A    & N/A   & N/A \\
\hline
\end{tabular}
\caption{Evaluation metrics for synthesis variants generated from the Meta vs. FTC scenario. All values are based on Sentence-BERT cosine similarity measures. “N/A” indicates that the synthesis length was too short for reliable scoring.}
\label{tab:evaluation}
\end{table}

\subsection*{Coverage Metric: Creative Synthesis vs. Literal Recombination}
The zero coverage scores reflect the design philosophy of our framework. We deliberately prioritize \textit{creative synthesis} — emergent strategic insights — over \textit{literal recombination}, aligning with computational creativity research on emergent generation.

\textbf{Illustrative Cases:}
\begin{itemize}[leftmargin=*,nosep]
\item \textit{``Meta's regulatory chess game...''} (Coverage: 0.00) integrates Machiavelli's maneuvering and Martin's stakeholder logic through metaphor
\item \textit{``Preserve the mask...''} demonstrates aphoristic distillation beyond literal recombination
\end{itemize}

\textbf{Forward Path:}
\begin{itemize}[leftmargin=*,nosep]
\item Metric extensions: thematic realization (metaphor) and creative emergence
\item Dynamic $\kappa_{ij}$ tuning for regulatory contexts
\item Action-aware coefficient schemes with expert validation
\end{itemize}

\subsection*{Implications and Next Steps}

This preliminary evaluation supports our hypothesis that entangled inference enables rich, expressive synthesis that cannot be reduced to heuristic selection. It also highlights several promising directions for future methodological refinement, not to correct apparent deficits, but to more fully capture the strengths of compositional reasoning:

\begin{itemize}
    \item \textbf{Threshold Calibration:} Adjusting similarity thresholds to better detect metaphorical or abstract heuristic realizations within creative synthesis.
    \item \textbf{Human Feedback Loops:} Incorporating qualitative ratings from expert judges to assess perceived relevance, insight, and persuasiveness beyond literal traceability.
    \item \textbf{Baseline Comparisons:} Evaluating entangled synthesis against simpler rule-ranking systems to clarify its added value in coherence, novelty, and strategic depth.
\end{itemize}

These enhancements can help align evaluation practices with the generative character of the system, especially in cases where literal coverage metrics underrepresent conceptual integration. More broadly, our findings contribute to ongoing discussions about emergent versus extractive generation in AI systems—an inquiry we consider foundational, although beyond the immediate scope of this study.

\paragraph{Note on Coverage Tuning.} While our system emphasizes creative synthesis over literal heuristic reuse, the coverage metric can be straightforwardly adapted for traceability-focused applications. For instance, a stricter prompting template could instruct the LLM to cite or paraphrase input heuristics explicitly, or a hybrid output mode could annotate generated narratives with back-references to contributing axioms. Such adaptations would increase literal coverage scores while reducing generativity, and may be appropriate in use cases requiring auditability or pedagogical transparency.

\section{Related Work}
\label{sec:related}
Our architecture intersects four major research traditions: quantum cognition, heuristic reasoning under uncertainty, semantic vector models of conceptual representation, and compositional reasoning.

\subsection*{Quantum Cognition and Non-Classical Reasoning}

A growing body of research has explored quantum models of cognition to explain violations of classical logic and probability in human decision-making. These models represent cognitive states as superpositions and decisions as contextual measurements, producing interference and entanglement effects analogous to those observed in quantum systems \cite{busemeyer2012quantum, busemeyer2011quantum, aerts2004quantum, aerts2013concepts}. Our entanglement logic draws from this tradition, treating heuristic interaction as a non-classical, context-sensitive process shaped by constructive or destructive interference rather than simple rule aggregation. Quantum cognition models have been increasingly applied to decision-making contexts \cite{pothos2022}, though critical perspectives—such as \cite{Lee_Vanpaemel_2013}—warn against overextending formal metaphors. This motivates our restrained use of the term ``entanglement,'' which we adopt as a semantic approximation, not as a psychological claim grounded in quantum mechanics.

\subsection*{Heuristics and Decision-Making under Uncertainty}

The psychological and computational power of simple heuristics in complex environments is well documented \cite{gigerenzer1999simple, hogarth2005simple}. Foundational work by Kahneman and Tversky \cite{kahneman1979prospect} highlighted how heuristics shape human judgment under uncertainty, often producing systematic deviations from classical rationality. While many systems focus on selecting the single best rule, our framework supports simultaneous activation and synthesis of multiple heuristics, even when their prescriptions conflict. This goes beyond bounded rationality models by emphasizing the creative and combinatorial aspects of strategic reasoning.

\subsection*{Semantic Representation and Cognitive Geometry}

We leverage advances in distributional semantics to encode axioms and scenarios as vectors in a high-dimensional latent space. This approach is grounded in word embedding models such as Word2Vec \cite{mikolov2013efficient} and extended through topic-based clustering for thematic organization \cite{rehurek2010framework, griffiths2007topics}. Our projection-based inference and similarity-weighted synthesis also resonate with quantum semantic models, where meaning is distributed and interaction-sensitive \cite{aerts2004quantum, aerts2013concepts}.

\subsection*{Conceptual Blending and Compositional Reasoning}

The heuristic synthesis process builds upon the theory of conceptual blending \cite{fauconnier2002way}, treating heuristics as partial conceptual spaces that can be combined into new, emergent forms. Our \texttt{mix($H_i$, $H_j$)} operator parallels blending operations in computational creativity systems \cite{veale2016metaphor}, but is guided by semantic vector similarity and strategic interference patterns.

\subsection*{Narrative Generation and Few-Shot Prompting}

Finally, our approach to LLM-based strategy generation connects with recent advances in prompt engineering and few-shot learning for generative dialogue systems \cite{madotto2020language}. Like these systems, our model uses a modular, structured prompt to elicit tailored outputs—but with a stronger emphasis on heuristic composition, rhetorical framing, and strategic intent.

\section{Conclusion and Future Work}
\label{sec:conclusion}

This paper presents a hybrid architecture for agent-augmented strategic reasoning, grounded in the extraction, activation, and synthesis of heuristics from classical and contemporary sources. Our central innovation lies in shifting from rule selection to heuristic composition, enabled by a quantum-inspired model of semantic entanglement and realized through an LLM-mediated synthesis engine.

Through both theoretical exposition and empirical evaluation—most notably, the Meta vs. FTC case study—we demonstrate that entangled synthesis enables context-sensitive strategy generation, integrating multiple, potentially conflicting heuristics into coherent and expressive formulations.

\subsection{Future Work: Methodological Priorities}

Three methodological directions emerge as immediate next steps:

\textbf{1. Context-Dependent Interference Modeling:}  
The current use of static $\kappa_{ij}$ coefficients limits contextual sensitivity. Future work should develop methods for:
\begin{itemize}
\item Learning context-specific interference patterns from expert judgments
\item Modeling how situational factors (e.g., regulatory pressure, crisis urgency) influence heuristic compatibility
\item Validating context-dependent models against real-world strategic cases
\end{itemize}

\textbf{2. Interference Coefficient Optimization:}  
Moving beyond uniform or similarity-based $\kappa_{ij}$ requires:
\begin{itemize}
\item Sensitivity analysis across coefficient configurations
\item Expert validation of heuristic compatibility ratings
\item Principled, potentially learned, approaches to coefficient determination
\item Investigation into whether optimal interference structures are domain-general or context-specific
\end{itemize}

\textbf{3. Evaluation Framework Enhancement:}  
Current metrics (coverage, coherence, novelty) remain incomplete proxies for synthesis quality. Future work should explore:
\begin{itemize}
\item Expert evaluation studies focused on insight, persuasiveness, and strategic coherence
\item Multi-dimensional metrics combining logical faithfulness, contextual creativity, and rhetorical framing
\item Comparative validation against historical and contemporary organizational strategies
\end{itemize}

\subsection{Deferred Priorities for Systematic Investigation}

Several critical directions have been intentionally deferred due to the exploratory nature of this implementation and the short revision timeline. These constitute long-term priorities for validating and extending the framework:

\begin{itemize}
    \item \textbf{Multiple Case Studies:} Statistical generalization across domains (e.g., defense, finance, public health) requires a broader case library and scenario corpus to assess transferability of synthesis patterns.
    
    \item \textbf{Expert Human Evaluation:} Strategic usefulness must be assessed through domain-expert judgment, ideally via blind comparisons of human- and machine-generated strategies.
    
    \item \textbf{Systematic Coefficient Learning:} Deriving $\kappa_{ij}$ values via supervised learning, expert annotation, or unsupervised semantic clustering presents a significant but essential methodological challenge.
    
    \item \textbf{Alternative Evaluation Frameworks:} Developing new metrics for strategic synthesis quality—capturing metaphor, abstraction, and integrative reasoning—will require interdisciplinary collaboration bridging AI, decision science, and strategy.
\end{itemize}

These directions define the roadmap for future iterations and offer multiple entry points for methodological innovation and interdisciplinary partnership.

\subsection{Methodological Limitations}

The current implementation makes several simplifying assumptions that should be revisited in future work:

\begin{enumerate}
\item \textbf{Coefficient Arbitrariness:} Static or uniform $\kappa_{ij}$ values lack empirical grounding
\item \textbf{Context Independence:} Interference modeling does not yet respond to scenario-specific variation
\item \textbf{Evaluation Gaps:} Current metrics may undervalue creative synthesis and metaphorical abstraction
\item \textbf{Selection Bias:} The strategist library may not reflect the full di
\end{enumerate}

We regard these limitations not as flaws of the approach but as invitations for systematic refinement.

\vspace{1em}
\noindent
We believe this architecture opens new avenues for human–AI collaboration in high-stakes domains, where the interplay of conceptual synthesis, narrative framing, and heuristic reasoning remains indispensable to strategic insight. In policy, finance, and organizational decision-making, such synthesis tools could assist leaders in navigating trade-offs among competing imperatives. However, real-time deployment in these contexts will require interpretability safeguards and institutional alignment.

\section*{Acknowledgments and Funding}
We thank Herv\'e Gallaire and the HAR 2025 reviewers, whose comments helped improve this article.

Remo Pareschi has been funded by the European Union--NextGenerationEU under the Italian Ministry of University and Research (MUR) National Innovation Ecosystem grant ECS00000041-VITALITY--CUP E13C22001060006.

\bibliographystyle{plain}
\bibliography{references}

\end{document}